\documentclass[11pt,letterpaper]{article}
\usepackage[top=0.85in,left=1.2in,footskip=0.75in,marginparwidth=2in]{geometry}

\usepackage[utf8]{inputenc}

\usepackage{cite}

\usepackage{nameref,hyperref}

\usepackage{amsmath}
\usepackage{float}
\usepackage[right]{lineno}
\usepackage{microtype}
\DisableLigatures[f]{encoding = *, family = * }

\raggedright
\usepackage{changepage}

\usepackage[aboveskip=1pt,labelfont=bf,labelsep=period,singlelinecheck=off]{caption}
\makeatletter
\renewcommand{\@biblabel}[1]{\quad#1.}
\makeatother

\usepackage{lastpage,fancyhdr,graphicx}
\usepackage{epstopdf}
\fancyheadoffset[L]{2.25in}
\fancyfootoffset[L]{2.25in}

\usepackage{color}
\definecolor{Gray}{gray}{.25}

\usepackage{graphicx}
\usepackage{sidecap}

\usepackage{wrapfig}
\usepackage[pscoord]{eso-pic}
\usepackage[fulladjust]{marginnote}
\reversemarginpar

\begin{document}
\vspace*{0.35in}

\begin{flushleft}
{\Large
\textbf\newline{DeepDTA: Deep Drug-Target Binding Affinity Prediction}
}
\newline
\\
Hakime \"{O}zt\"{u}rk\textsuperscript{1,*},
Elif Ozkirimli\textsuperscript{2,*},
and Arzucan \"{O}zg\"{u}r\textsuperscript{1,*}

\bigskip
\bf{1} Department of Computer Engineering, Bogazici University, Istanbul, 34342, Turkey
\\
\bf{2} Department of Chemical Engineering, Bogazici University, Istanbul, 34342, Turkey
\\
\bigskip
* hakime.ozturk@boun.edu.tr; elif.ozkirimli@boun.edu.tr; arzucan.ozgur@boun.edu.tr

\end{flushleft}
\justify
\section*{Abstract}
The identification of novel drug-target (DT) interactions is a substantial part of the drug discovery process. Most of the computational methods that have been proposed to predict DT interactions have focused on binary classification, where the goal is to determine whether a DT pair interacts or not. However, protein-ligand interactions assume a continuum of binding strength values, also called binding affinity and predicting this value still remains a challenge. The increase in the affinity data available in DT knowledge-bases allows the use of advanced learning techniques such as deep learning architectures in the prediction of binding affinities. 
In this study, we propose a deep-learning based model that uses only sequence information of both targets and drugs to predict DT interaction binding affinities. The few studies that focus on DT binding affinity prediction use either 3D structures of protein-ligand complexes or 2D features of compounds. One novel approach used in this work is the modeling of protein sequences and compound 1D representations with convolutional neural networks (CNNs).
The results show that the proposed deep learning based model that uses the 1D representations of targets and drugs is an effective approach for drug target binding affinity prediction. The model in which  high-level representations of a drug and a target are constructed via CNNs achieved the best Concordance Index (CI) performance in one of our larger benchmark data sets, outperforming the KronRLS algorithm and SimBoost, a  state-of-the-art method for DT binding affinity prediction.

\section*{Introduction}

The successful identification of drug-target interactions (DTI) is a critical step in drug discovery. As the field of drug discovery expands with the discovery of new drugs, repurposing of existing drugs and identification of novel interacting partners for approved drugs is also gaining interest \cite{oprea2012drug}. Until recently, DTI prediction was approached as a binary classification problem \cite{yamanishi2008predict, bleakley2009, laarhoven2011, gonen2012predict, cao2014computational, cao2012large, cobanoglu2013predict, ozturk2016comparative}, neglecting an important piece of information about protein-ligand interactions, namely the binding affinity values. Binding affinity provides information on the strength of the interaction between a drug-target (DT) pair and it is usually expressed in measures such as dissociation constant ($K_d$), inhibition constant ($K_i$), or the half maximal inhibitory concentration (IC50). IC50 depends on the concentration of the target and ligand \cite{cer2009ic} and low IC50 values signal strong binding. 
Similarly, low $K_i$ values indicate high binding affinity. $K_d$ and $K_i$ values are usually represented in terms of p$K_d$ or p$K_i$, the negative logarithm of the dissociation or inhibition constants. 

In binary classification based DTI prediction studies, construction of the data sets constitutes a major problem, since negative (not-binding) information is generally hard to find. In most cases, the DT pairs for which binding information is not known are treated as negative (not-binding) samples.  The lack of true-negative samples and how the study generates synthetic negative samples usually affects the performance of the prediction algorithms. On the other hand, formulating the DT prediction task as a binding affinity prediction problem enables the creation of more realistic data sets, where the binding affinity scores are directly used, obviating the need for the generation of synthetic negative samples. 

Prediction of protein-ligand  binding affinities has been the focus of protein-ligand scoring, which is frequently used after virtual screening and docking campaigns in order to predict the putative strengths of the proposed ligands to the target \cite{ragoza2017protein}. Non-parametric machine learning methods such as the Random Forest (RF) algorithm have been used as a successful alternative to scoring functions that depend on multiple parameters  \cite{ballester2010machine,li2015low,shar2016pred}. However, Gabel et al. showed that RF-score failed in virtual screening and docking tests, speculating that using features such as co-occurrence of atom-pairs over-simplified the description of the protein-ligand complex and led to the loss of information that the raw interaction complex could provide \cite{gabel2014beware}. Around the same time this study was published, deep learning started to become a popular architecture powered by the increase in data and high capacity computing machines challenging machine learning methods. 

Inspired by the remarkable success rate in image processing \cite{ciregan2012multi, donahue2014decaf, simonyan2014very} and speech recognition \cite{hinton2012deepspeech, dahl2012context, graves2013speech}, deep learning methods are now being intensively used in many other research fields, including bioinformatics such as in genomics studies \cite{leung2014deep, xiong2015human} and quantitative-structure activity relationship (QSAR) studies in drug discovery \cite{ma2015deep}. The major advantage of deep learning architectures is that they enable better representations of the raw data by non-linear transformations in each layer \cite{lecun2015deep} and thus they facilitate learning the hidden patterns in the data. 

A few studies employing Deep Neural Networks (DNN) have already been performed for DTI binary class prediction using different input models for proteins and drugs \cite{tian2015boosting, chan2016large, hamanaka2016cgbvs} in addition to some studies that employ stacked auto-encoders \cite{wang2017computational} and deep-belief networks \cite{wen2017deep}. Similarly, stacked auto-encoder based models with Recurrent Neural Networks (RNNs) and Convolutional Neural Networks (CNNs) were applied to represent chemical and genomic structures in real-valued vector forms \cite{gomez2016automatic, jastrzkebski2016learning}.  Deep learning approaches have also been applied to protein-ligand interaction scoring in which  a common application has been the use of CNNs that learn from the 3D structures of the protein-ligand complexes \cite{wallach2015atomnet, ragoza2017protein, gomes2017atomic}. However, this approach is limited to known protein-ligand complex structures, with only 25000 ligands reported in PDB \cite{rose2016rcsb}. 

Pahikkala et al. employed the Kronecker Regularized Least Squares (KronRLS) algorithm that  utilizes only 2D based compound similarity-based representations of the drugs and Smith-Waterman similarity representation of the targets \cite{pahikkala2014toward}.  Recently, SimBoost method was  proposed to predict binding affinity scores with a gradient boosting machine by using feature engineering to represent drug-target interactions \cite{he2017simboost}. They utilized similarity-based information of DT pairs as well as features that were extracted from network-based interactions between the pairs. Both studies used traditional machine learning algorithms and utilized 2D-representations of the compounds in order to obtain similarity information.

In this study, we propose an approach to predict the binding affinities of protein-ligand interactions with deep learning models using only  sequences (1D representations) of proteins and ligands. To this end, the sequences of the proteins and SMILES (Simplified Molecular Input Line Entry System) representations of the compounds are used rather than external features or 3D-structures of the binding complexes. We employ CNN blocks to learn representations from the raw protein sequences and SMILES strings and combine these representations to feed into a fully-connected layer block that we call DeepDTA. We use the Davis Kinase binding affinity data set \cite{davis2011comprehensive} and the KIBA large-scale kinase inhibitors bioactivity data \cite{tang2014making, he2017simboost} to evaluate the performance of our model and compare our results with  the KronRLS \cite{pahikkala2014toward} and SimBoost algorithms  \cite{he2017simboost}.  Our new model that uses two separate CNN-based blocks to represent proteins and drugs performs as well as the KronRLS  and SimBoost algorithms on the Davis data set, and  it performs significantly better than both the KronRLS and SimBoost algorithms  on the KIBA data (p-value, 0.0001).  With our proposed model, we also obtain the lowest Mean Squared Error (MSE) value on both data sets.

\section*{Materials and Methods}

\subsection*{Data sets}
We evaluated our proposed model on two different data sets, the Kinase data set Davis \cite{davis2011comprehensive}  and KIBA data set \cite{tang2014making}, which were previously used as benchmark data sets for binding affinity prediction evaluation \cite{pahikkala2014toward,he2017simboost}. 

The Davis data set contains selectivity assays of the kinase protein family and the relevant inhibitors with their respective dissociation constant ($K_d$) values. It comprises interactions of 442 proteins and 68 ligands. The KIBA data set, on the other hand, originated from an approach called KIBA, in which kinase inhibitor bioactivities from different sources such as $K_i$, $K_d$, and $IC_{50}$ were combined \cite{tang2014making}.  KIBA scores were constructed to optimize the consistency between $K_i$, $K_d$, and $IC_{50}$ by utilizing the statistical information they contained. The KIBA data set originally comprised 467 targets and 52498 drugs. \cite{he2017simboost}  filtered it to contain only drugs and targets with at least 10 interactions yielding  a total of 229 unique proteins and 2111 unique drugs. Table \ref{Tab:01} summarizes these data sets in the forms that we used in our experiments.

\begin{table}[h]
\caption{Data set} \label{Tab:01} {
\begin{tabular}{@{}llll@{}}\hline
& Proteins & Compounds & Interactions\\\hline
Davis ($K_d$) & 442 & 68 & 30056 \\ \hline
KIBA  & 229 & 2111 & 118254 \\ \hline
\end{tabular}}
\end{table}

\begin{figure*}[htp!]
\centerline{\includegraphics[scale=0.9]{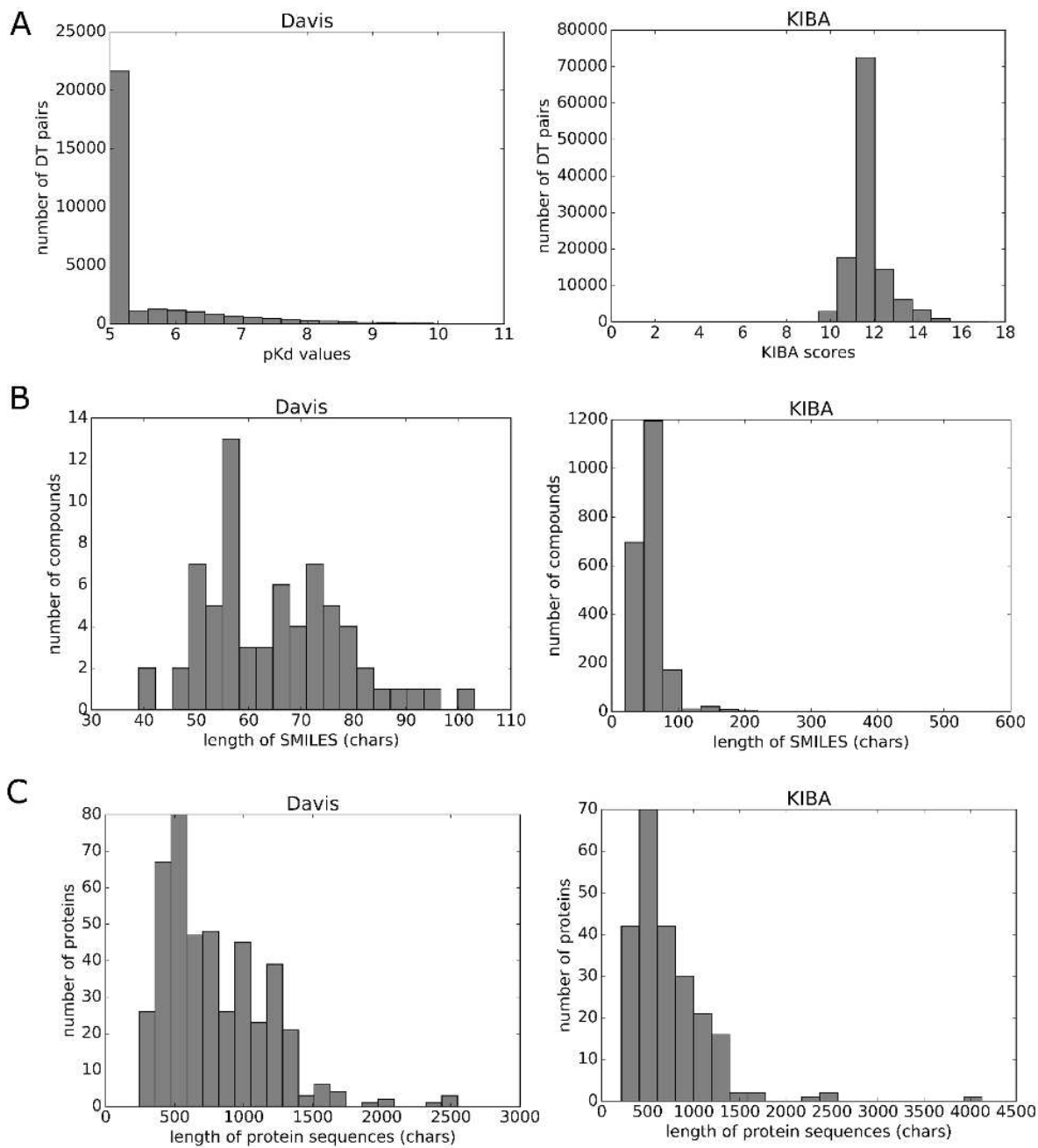}}
\caption{Summary of the Davis (left panel) and KIBA  (right panel) data sets. A) Distribution of binding affinity values B) Distribution of the lengths of the SMILES strings C) Distribution of the lengths of the protein sequences .} \label{fig:01}
\end{figure*}

While \cite{pahikkala2014toward} used the $K_d$ values of the Davis data set directly as the binding affinity values, we used the values transformed into log space,  $pK_d$, similar to \cite{he2017simboost} as explained in Equation \ref{pkd}.
\begin{equation}\label{pkd}
pK_d= -log10 (\frac{K_d}{1e9})
\end{equation}
Figure \ref{fig:01}A (left panel) illustrates the distribution of the binding affinity values in $pK_d$  form. The peak at $pK_d$ value 5 (10000$nM$)  constitutes more than half of the data set (20931 out of 30056). These values correspond to the negative pairs that either have very weak binding affinities ($K_d > 10000nM$) or are not observed in the primary screen \cite{pahikkala2014toward}. As such they are true negatives.

The distribution of the KIBA scores is depicted in the right panel of Figure \ref{fig:01}A. \cite{he2017simboost} pre-processed the KIBA scores as follows: (i)  for each KIBA score, its negative was taken, (ii) the minimum value among the negatives was chosen, and (iii) the  absolute value of the minimum was added to all negative scores, thus constructing the final form of the KIBA scores. 

The compound SMILES strings of the Davis data set were extracted from the Pubchem compound database based on their Pubchem CIDs \cite{bolton2008pubchem}.  For KIBA, first the CHEMBL IDs were converted into Pubchem CIDs and  then, the corresponding CIDs were used to extract the SMILES strings. Figure \ref{fig:01}B  illustrates the distribution of the lengths of the SMILES strings of the compounds in the Davis (left) and KIBA (right) data sets. For the compounds of the Davis dataset, the maximum length of a SMILES is 103, while the average length is equal to 64.  For the compounds of KIBA, the maximum length of a SMILES is 590, while the average length is equal to 58.

The protein sequences of the Davis data set were extracted from the UniProt protein database based on gene names/RefSeq accession numbers  \cite{apweiler2004uniprot}. Similarly, the UniProt IDs of the targets in the KIBA data set were used to collect the protein sequences. Figure \ref{fig:01}C left panel shows the lengths of the sequences of the proteins in the Davis data set. The maximum length of a protein sequence is 2549 and the average length is 788 characters. Figure \ref{fig:01}C right panel depicts the distribution of protein sequence length in KIBA targets.  The maximum length of a protein sequence is 4128 and the average length is 728 characters.

We should also note that the Smith-Waterman (S-W) similarity among proteins of the KIBA data set is at most 60\%  for 99\% of the protein pairs. The target similarity is at most 60\% for 92\% of the protein pairs for the Davis data set. These statistics indicate that both data sets are  non-redundant.

\subsection*{Input Representation}

We used  integer/label encoding that  uses integers  for the categories  to  represent inputs. We scanned  approximately 2M SMILES sequences that we collected from Pubchem and compiled 64 labels (unique letters). For protein sequences, we scanned 550K protein sequences from UniProt and extracted 25 categories (unique letters). 

Here we  represent each label with a corresponding integer (e.g. ``C'':1, ``H'':2, `N'':3 etc.). The label encoding for the example SMILES, ``CN=C=O'', is given below.  

 \begin{align*}
   \begin{bmatrix}
           C  & N & = & C & = & O
         \end{bmatrix}
&= 
\begin{bmatrix}
		1 & 3 & 63 & 1 & 63 & 5
         \end{bmatrix}
\end{align*}

Protein sequences are encoded in a similar way using label encodings. Both SMILES and protein sequences have varying lengths. Hence, in order to create an effective representation form, we decided on fixed maximum lengths of 85 for SMILES and 1200 for protein sequences for Davis. To represent the components of KIBA, we chose the maximum 100 characters length for SMILES and 1000 for protein sequences. We chose these maximum lengths based on the distributions illustrated in Figure \ref{fig:01}B and \ref{fig:01}C so that the maximum lengths cover at least 80\% of the proteins and 90\% of the compounds in the data sets. The sequences that are longer than the maximum length are truncated, whereas shorter sequences are 0-padded.

\subsection*{Proposed Model }
\label{s:proposed}
In this study we treated protein-ligand interaction prediction as a regression problem by aiming to predict the binding affinity scores.  As a prediction model, we  adopted a  popular  deep learning architecture, Convolutional Neural Network (CNN).  CNN is an architecture that contains one or more convolutional layers often followed by a pooling layer. A pooling layer down-samples the output of the previous layer and provides a way of generalization of the features that are learned by the filters.  On  top of the convolutional and pooling layers, the model is completed with one or more fully connected  (FC) layers. The most powerful feature of  CNN models is their ability to capture the local dependencies with the help of filters. Therefore, the number and size of the filters in a CNN directly affects the type of features the model learns from the input. It is often reported that as the number of filters increases, the model becomes better at recognizing patterns \cite{kang2014convolutional}.

\begin{figure}[H]
\centerline{\includegraphics[scale=0.18]{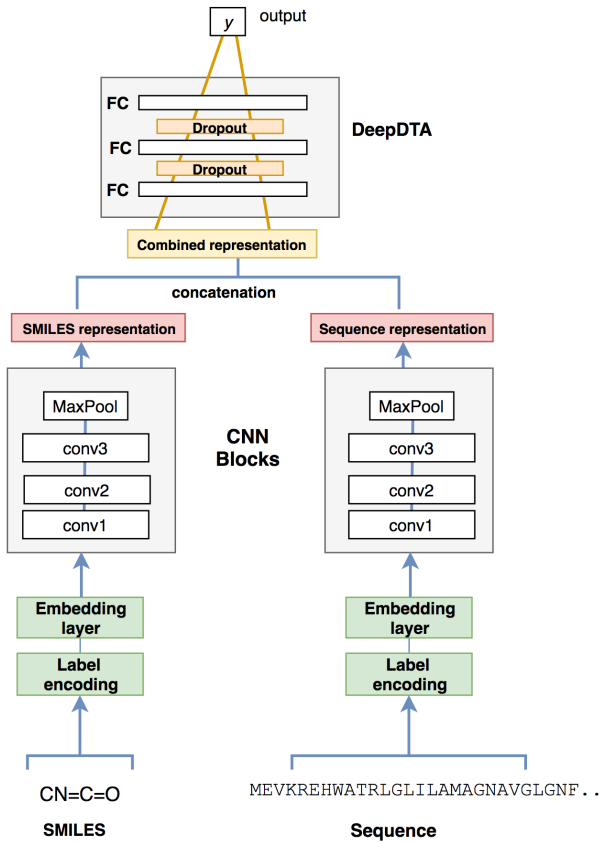}}
\caption{DeepDTA model with two CNN blocks to learn from compound SMILES and protein sequences.}\label{fig:03}
\end{figure}

We proposed a CNN-based prediction model  that comprises two separate CNN blocks, each of which aims to learn representations from SMILES strings and protein sequences. For each CNN block, we used three consecutive 1D-convolutional layers with increasing number of filters. The second layer had double and the third convolutional layer had triple the number of filters in the first one. The convolutional layers were then followed by the max-pooling layer. The final features of the max-pooling layers were concatenated and fed into three FC layers, which we named as DeepDTA. We used 1024 nodes in the first two FC layers, each followed by a dropout layer of rate 0.1. Dropout  is  a regularization technique that is used to avoid over-fitting by setting the activation of some of the neurons to 0  \cite{srivastava2014dropout}.  The third layer consisted of 512 nodes and was followed by the output layer. The proposed model that combines two CNN blocks is illustrated in Figure \ref{fig:03}.

As the activation function, we used Rectified Linear Unit (ReLU) \cite{nair2010rectified}, $g(x) = max(0, x)$,  which has been widely used in deep learning studies \cite{lecun2015deep}. A learning model tries to minimize the difference between the expected (real) value and the prediction during training. Since we work on a regression task, we used mean squared error (MSE) as the loss function, in which $P$ is the prediction vector, and $Y$ corresponds to the vector of actual outputs. $n$ indicates the number of samples.
\begin{equation}
MSE= \dfrac{1}{n}\sum_{i=1}^{n}(P_i - Y_i)^2
\end{equation}
The learning  was completed with 100 epochs and mini-batch size of 256 was used to update the weights of the network. Adam was used as the optimization algorithm to train the networks \cite{kingma2014adam} with the default learning rate  of 0.001.
We used Keras' Embedding layer to represent characters with 128-dimensional dense vectors. The input for Davis data set consisted of (85,128) and (1200, 128) dimensional matrices for the compounds and proteins, respectively. We represented KIBA data set with a (100,128) dimensional matrix for the compounds and a (1000, 128) dimensional matrix for the  proteins.

\section*{Experiments and Results}

Here, we propose a novel drug - target binding affinity prediction method based on only sequence information of compounds and proteins. We utilized the Concordance Index (CI) to measure the performance of the proposed model  and compared it with the current state-of-art methods that we chose as our baselines, namely a Kronecker Regularized Least Squares (KronRLS) based approach \cite{pahikkala2014toward} and SimBoost \cite{he2017simboost}.  We provide more information about these baseline methodologies, our model and experimental setup, as well as our results in the following subsections.

\subsection*{Baselines}

\subsubsection*{Kron-RLS}
KronRLS  aims to minimize the following function, where $f$ is the prediction function \cite{pahikkala2014toward}:
\begin{equation}\label{e:kronls}
J(f) = \sum_{i=1}^m {(y_i - f(x_i) )^2} + \lambda ||f||^2_k
\end{equation}

$||f||^2_k$ is the norm of $f$, which is related to the kernel function $k$, and
$\lambda>0$ is a regularization hyper-parameter defined by the user.  A minimizer for Equation \ref{e:kronls} can be defined as follows \cite{kimeldorf1971some}:
\begin{equation}\label{e:minimizer}
f(x) = \sum_{i=1}^m a_i k(x, x_i)
\end{equation}
where $k$ is the kernel function. In order to represent compounds, they utilized a similarity matrix   computed using Pubchem structure clustering server (Pubchem Sim)(http://pubchem.ncbi.nlm.nih.gov), a tool that utilizes single linkage for cluster and uses  2D properties of the compounds to measure their similarity. As for proteins, the Smith-Waterman algorithm was used to construct a protein similarity matrix \cite{smith1981identification}.

\subsubsection*{SimBoost}

SimBoost is a gradient boosting machine based method that depends on the features constructed from drugs, targets and drug-target pairs \cite{he2017simboost}.  The proposed methodology uses feature engineering to build three types of features: (i) object-based features that utilize occurrence statistics and pairwise similarity information of drugs and targets, (ii) network-based features such as neighbor statistics, network metrics (betweenness, closeness etc.), PageRank score, which are collected from the respective drug-drug and target-target networks (In a drug-drug network, drugs are represented as nodes and connected to each other if the similarity of these two drugs is above a user-defined threshold. The target-target network is constructed in a similar way.) (iii) network-based features that are collected from a heterogeneous network (drug-target network) where a node can either be a drug or target and the drug nodes and target nodes  are connected to each other via binding affinity value. In addition to the network metrics, neighbor statistics and PageRank scores, as well as latent vectors from matrix factorization are also included in this type of network.


These features are fed into a supervised learning method named gradient boosting regression trees \cite{chen2015higgs,chen2016xgboost} derived from gradient boosting machine model \cite{friedman2001greedy}. With gradient boosting regression trees, for a given drug-target pair $dt_i$, the binding affinity score $\bar{y}$ predicted as follows \cite{he2017simboost}:

\begin{equation}\label{e:simboost}
\bar{y_i} =  \theta(dt_i)=\sum_{m=1}^M {f_m(dt_i)}, f_m\in F
\end{equation}
in which $M$ denotes the number of regression trees and $F$ represents the space of all possible trees.  A regularized objective function to learn the set of trees ${f_m}$ is described in the following form \cite{he2017simboost}:
\begin{equation}\label{e:simboosto}
R(\theta) = \sum_{i} {l(y_i, \bar{y}_i)} +  \sum_{m} \alpha (f_m)
\end{equation}
where $l$ is the loss function that measures the difference between the actual binding affinity value $y_i$ and the predicted value $\bar{y}_i$, while $\alpha$ is the tuning parameter that controls the complexity of the model. The details are described in \cite{he2017simboost,chen2015higgs,chen2016xgboost}. Similar to  \cite{pahikkala2014toward}, \cite{he2017simboost} also used PubChem clustering server for drug similarity and Smith-Waterman for protein similarity computation.

\subsection*{Evaluation Metrics}

To evaluate the performance of a model that outputs continuous values, Concordance Index (CI) was used \cite{gonen2005concordance}:
\begin{equation}
CI = \frac {1} {Z} \sum_{\delta_i > \delta_j} h (b_i - b_j)
\end{equation}
where $b_i$ is the prediction value for the larger affinity $\delta_i$, $b_j$ is the prediction value for the smaller affinity $\delta_j$, $Z$ is a normalization constant, $h(m)$ is the step function \cite{pahikkala2014toward}:
\begin{equation}
   h(x) = 
\begin{cases} 
          1,  & \text{if}\ x > 0\\
         0.5,  & \text{if}\ x= 0\\
	   0, & \text{if}\ x <0 
 \end{cases}                                          
 \end{equation}
The metric measures whether the predicted binding affinity values of two random drug-target pairs were predicted in the same order as their true values were. We used paired-t test for the statistical significance tests with 95\% confidence interval. We also used MSE, which was explained in Section \ref{s:proposed}, as an evaluation metric.

\subsection*{Experiment Setup}

\begin{figure}[h]
\centerline{\includegraphics[scale=0.1]{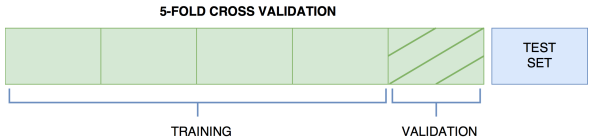}}
\caption{Experiment setup.}\label{fig:04}
\end{figure}

We evaluated the performance of the proposed model on the benchmark data sets \cite{davis2011comprehensive,tang2014making} similarly to \cite{he2017simboost}. They used nested-cross validation to decide on the best parameters for each test set. In order to learn a generalized model, we randomly divided our data set into six equal parts in which one part is selected as the independent test set. The remaining parts of the data set were used to determine the hyper-parameters via five-fold cross validation. Figure \ref{fig:04} illustrates the partitioning of the data set. The same setting with the same train and test folds was used for  KronRLS \cite{pahikkala2014toward} and Simboost \cite{he2017simboost} for a fair comparison.

We decided on three hyper-parameters for our model,  namely the number of the filters (same for proteins and compounds), the length of the filter size for compounds, and the length of the filter size for proteins.  We opted to experiment with different filter lengths for compounds and proteins instead of a common length, due to the fact that they have different alphabets. The  hyper-parameter combination that provided the best average CI score over the validation set was selected as the best combination in order to model the test set.  We first experimented with hyper-parameters chosen from a wide range and then fine-tuned the model. For example, to determine the number of filters we performed a search over [16, 32, 64, 128, 512].  We then narrowed the search range around the best performing parameter (e.g. if 16 was chosen as the best parameter, then our range was updated as [4, 8, 16, 20] etc.). 

As explained in the Proposed Model subsection, the second convolution layer was set to contain twice the number of filters of the first layer, and the third one was set to contain three times the number of filters of the first layer. 32 filters gave the best results over the cross-validation experiments. Therefore, in the final model, each CNN block consisted of three 1D convolutions of 32, 64, 96 filters. For all test results reported in Table \ref{tab:02} we used the same structure summarized in Table \ref{tab:paramss1} except for the lengths of the pre-fine-tuned filters that were used for the compound CNN-block and protein CNN-block.

\begin{table}[H]
\centering
\caption{Parameter settings for CNN based DeepDTA model}
\label{tab:paramss1}
\begin{tabular}{ll}
\hline
\textbf{Parameters} & \textbf{Range}  \\ \hline
Number of filters      & 32*1; 32*2; 32*3   \\ \hline
Filter length (compounds)        &   [4,6,8]  \\ \hline
Filter length (proteins) & [4,8,12]  \\ \hline
epoch & 100  \\ \hline
hidden neurons & 1024; 1024; 512  \\ \hline
batch size & 256   \\ \hline
dropout & 0.1 \\ \hline
optimizer & Adam \\ \hline
learning rate (lr) & 0.001  \\ \hline

\end{tabular}
\end{table}

In order to provide a more robust performance measure, we evaluated the performance over the independent test set, when the model was trained with the learned parameters in Table \ref{tab:paramss1} on the five training sets that we used in five-fold cross validation (note that the validation sets were not used).  The final CI score was reported as the average of these five results. Keras \cite{chollet2015keras} with Tensorflow \cite{abadi2016tensorflow} back-end was used as development framework. Our experiments were run on OpenSuse 13.2 (3.50GHz  Intel(R) Xeon(R) and GeForce GTX 1070  (8GB)). The work was accelerated by running on GPU with cuDNN \cite{chetlur2014cudnn}. We provide the train and test folds of the data sets (https://github.com/hkmztrk/DeepDTA/).

\subsection*{Results}

In this study, we propose a deep-learning model that uses two CNN-blocks to learn representations for drugs and targets based on their sequences. As a baseline for comparison, the KronRLS algorithm and SimBoost methods that use similarity matrices for proteins and compounds as input were used. The Smith-Waterman (S-W) and Pubchem Sim algorithms were used to compute the pairwise similarities for the proteins and ligands, respectively.  We then used these S-W and Pubchem Sim similarity scores as inputs to the FC part of our model (DeepDTA) to evaluate the model. Finally, we  used three alternative combinations in learning the hidden patterns of the data and used this information as input to our DeepDTA model. The combinations were $(i)$ learning only compound representation with a CNN block and using S-W similarity as protein representation , $(ii)$ learning only protein sequence representation with a CNN block and using Pubchem Sim to describe compounds, and (\textit{iii}) learning both protein representation and compound representations with a CNN block. We call the last combination used with DeepDTA the combined model.

Tables \ref{tab:02} and \ref{tab:03} report the average MSE and CI scores over the independent test set of the five models trained with the same parameters (shown in Table \ref{tab:paramss1}) using the five different training sets for Davis and KIBA data sets. 

\begin{table}[H]
\caption{The average CI and MSE scores of the test set trained on five different training sets for the Davis data set. The standard deviations are given in parenthesis. }  \label{tab:02} 
\scalebox{0.8}{
\begin{tabular}{@{}llllll@{}}\hline
 & Proteins & Compounds & CI  (std) & MSE\\\hline
KronRLS \cite{pahikkala2014toward} & Smith-Waterman & Pubchem Sim & 0.871 (0.0008)  & 0.379   \\

SimBoost \cite{he2017simboost} & Smith-Waterman & Pubchem Sim & 0.872 (0.002) & 0.282  \\

DeepDTA &  Smith-Waterman & Pubchem Sim & 0.790 (0.009)  & 0.608 \\

DeepDTA   & CNN & Pubchem Sim & 0.835 (0.005)    & 0.419  \\

DeepDTA  & Smith-Waterman & CNN & \textbf{0.886} (0.008)   &  0.420 \\

DeepDTA  & CNN & CNN & 0.878 (0.004)   &  \textbf{0.261} \\

\end{tabular}}
\end{table}

\begin{table}[H]
\caption{The average CI and MSE scores of the test set trained on five different training sets for the KIBA data set. The standard deviations are given in parenthesis.  }  \label{tab:03}  
\scalebox{0.8}{
\begin{tabular}{@{}llllll@{}}\hline
&  Proteins & Compounds & CI  (std) & MSE\\\hline
KronRLS \cite{pahikkala2014toward} & Smith-Waterman & Pubchem Sim &  0.782 (0.0009) & 0.411    \\

SimBoost \cite{he2017simboost} & Smith-Waterman & Pubchem Sim & 0.836 (0.001)  & 0.222  \\

DeepDTA &  Smith-Waterman & Pubchem Sim & 0.710 (0.002)  & 0.502  \\

DeepDTA  & CNN & Pubchem Sim & 0.718 (0.004)   & 0.571   \\

DeepDTA   & Smith-Waterman & CNN & 0.854 (0.001)  & 0.204  \\

DeepDTA  & CNN & CNN & \textbf{0.863} (0.002)  & \textbf{0.194}  \\

\end{tabular}}
\end{table}

In the Davis data set,  SimBoost and KronRLS methods perform similarly while the CI values for SimBoost is higher than that for KronRLS in the larger KIBA dataset. When the similarity measures S-W, for proteins, and Pubchem Sim, for compounds, are used with the the fully-connected part of the neural networks (DeepDTA), the CI drops to 0.79 for the Davis data set and to 0.71 for the KIBA data set. The MSE increases to more than 0.5. These results suggest that  the use of a feed-forward neural network with predefined features is not sufficient  to describe drug target interactions and to predict drug target affinities. Therefore, we used CNN layers to learn representations of drugs and proteins to capture hidden patterns in the datasets.

We first used CNN to learn representations of proteins and used the predefined Pubchem Sim scores for the ligands. Using this combination did not improve the results suggesting that use of a CNN architecture is not effective enough to learn from amino acid sequences. 

Then we used the CNN block to learn compound representations from SMILES and used the predefined S-W scores for the proteins. This combination outperformed the baselines on the KIBA data set with statistical significance (p-value of 0.0001 for both SimBoost and KronRLS), and on the Davis data set  (p-value of around 0.03 for both SimBoost and KronRLS). These results suggested that the CNN is able to capture more information than Pubchem Sim in the compound representation task.

Motivated by this result, we tested the combined CNN model in which both protein and compound representations are learned from the CNN layer. This method performed as well as the baseline methods with CI score of 0.878 on the Davis data set and achieved the best  CI score (0.863)  on the KIBA data set with statistical significance over both baselines (p-value of 0.0001 for both).  The MSE values of this model were also notably lower than the MSE of the baseline models on both data sets. Even though learning protein representations with CNN was not effective, combination of the two CNN blocks for proteins and ligands provided a strong model.

\section*{Conclusion}

We propose a deep-learning based approach to predict drug-target binding affinity using only sequences of proteins and drugs. We use Convolutional Neural Networks (CNN) to learn representations from the raw sequence data of proteins and drugs and  fully connected layers (DeepDTA) in the affinity prediction task. We compare the performance of the proposed model with two recent studies that employed the KronRLS regression algorithm \cite{pahikkala2014toward}  and the SimBoost method \cite{he2017simboost} as our baselines. We perform our experiments on the Davis kinase - drug data set and the KIBA data set. 

Our results showed that the use of predefined features with DeepDTA is not sufficient to describe protein - ligand interactions. However, when two CNN-blocks that learn representations of proteins and drugs based on raw sequence data are used in conjunction  with DeepDTA, the performance increases  significantly compared to both baseline methodologies for both KIBA and Davis data sets. Furthermore, the model that uses CNN to learn compound representations from SMILES and S-W similarities of proteins also achieves better performance than the baselines.

We  observed that the model that uses CNN-block to learn proteins and 2D compound similarity to represent compounds performed poorly compared to the other methods that employ CNN.  This might be an indication that amino-acids  require a structure that can handle their ordered relationships, which the CNN architecture failed to capture successfully. Long-Short Term Memory (LSTM), which is a special type of Recurrent Neural Networks (RNN), could be a more suitable approach to learn from protein sequences, since the architecture has memory blocks that allow effective learning from a long sequence. LSTM architecture has been successfully employed to tasks such as  detecting homology \cite{hochreiter2007fast}, constructive peptide design \cite{muller2018recurrent}  and function prediction \cite{liu2017deep} that utilize amino-acid sequences. As future work, we also aim to utilize a recent ligand-based protein representation method proposed by our team that uses SMILES sequences of the interacting ligands to describe proteins \cite{ozturk2018novel} .

The results indicated that deep-learning based methodologies performed notably better than the baseline methods with a statistical significance when the data set grows in size, as the KIBA data set is four times larger than the Davis data set. The improvement over the baseline was significantly higher for the KIBA data set (from CI score of 0.836 to 0.863) compared to the Davis data set (from CI score of 0.872 to 0.878). The increase in the data enables the deep learning architectures to capture the hidden information better. 

The major contribution of this study is the presentation of a novel deep learning-based model for drug - target affinity prediction that  uses only character representations of  proteins and drugs. By simply using raw sequence information for both drugs and targets, we were able to achieve similar or better performance than the baseline methods that depend on  multiple different tools and algorithms to extract features.   

A large percentage of proteins remains untargeted, either due to bias in the drug discovery field for a select group of proteins or due to their undruggability, and this untapped pool of proteins has gained interest with protein deorphanizing efforts \cite{edwards2011too, o2016ligand,fedorov2010targeted}. As future work, we will focus on building an effective representation for protein sequences. The methodology can then be extended to predict the affinity of known compounds/targets to novel targets/drugs as well as to the prediction of the affinity of  novel drug-target pairs.

\section*{Acknowledgments}
TUBITAK-BIDEB 2211-E Scholarship Program (to HO) and BAGEP Award of the Science Academy (to AO) are gratefully acknowledged.  We thank Ethem Alpaydın, Attila Gürsoy and Pınar Yolum for the helpful discussions.\vspace*{-12pt}

\section*{Funding}

This work is funded by Bogazici University Research Fund (BAP) Grant Number 12304.

\nolinenumbers

\bibliography{main}

\begin{thebibliography}{10}

\bibitem{abadi2016tensorflow}
M.~Abadi, A.~Agarwal, P.~Barham, E.~Brevdo, Z.~Chen, C.~Citro, G.~S. Corrado,
  A.~Davis, J.~Dean, M.~Devin, et~al.
\newblock Tensorflow: Large-scale machine learning on heterogeneous distributed
  systems.
\newblock {\em arXiv preprint arXiv:1603.04467}, 2016.

\bibitem{apweiler2004uniprot}
R.~Apweiler, A.~Bairoch, C.~H. Wu, W.~C. Barker, B.~Boeckmann, S.~Ferro,
  E.~Gasteiger, H.~Huang, R.~Lopez, M.~Magrane, et~al.
\newblock Uniprot: the universal protein knowledgebase.
\newblock {\em Nucleic acids research}, 32(suppl\_1):D115--D119, 2004.

\bibitem{ballester2010machine}
P.~J. Ballester and J.~B. Mitchell.
\newblock A machine learning approach to predicting protein--ligand binding
  affinity with applications to molecular docking.
\newblock {\em Bioinformatics}, 26(9):1169--1175, 2010.

\bibitem{bleakley2009}
K.~Bleakley and Y.~Yamanishi.
\newblock Supervised prediction of drug--target interactions using bipartite
  local models.
\newblock {\em Bioinformatics}, 25(18):2397--2403, 2009.

\bibitem{bolton2008pubchem}
E.~E. Bolton, Y.~Wang, P.~A. Thiessen, and S.~H. Bryant.
\newblock Pubchem: integrated platform of small molecules and biological
  activities.
\newblock {\em Annual reports in computational chemistry}, 4:217--241, 2008.

\bibitem{cao2012large}
D.-S. Cao, S.~Liu, Q.-S. Xu, H.-M. Lu, J.-H. Huang, Q.-N. Hu, and Y.-Z. Liang.
\newblock Large-scale prediction of drug--target interactions using protein
  sequences and drug topological structures.
\newblock {\em Analytica chimica acta}, 752:1--10, 2012.

\bibitem{cao2014computational}
D.-S. Cao, L.-X. Zhang, G.-S. Tan, Z.~Xiang, W.-B. Zeng, Q.-S. Xu, and A.~F.
  Chen.
\newblock Computational prediction of drug- target interactions using chemical,
  biological, and network features.
\newblock {\em Molecular Informatics}, 33(10):669--681, 2014.

\bibitem{cer2009ic}
R.~Z. Cer, U.~Mudunuri, R.~Stephens, and F.~J. Lebeda.
\newblock Ic 50-to-k i: a web-based tool for converting ic 50 to k i values for
  inhibitors of enzyme activity and ligand binding.
\newblock {\em Nucleic acids research}, 37(suppl\_2):W441--W445, 2009.

\bibitem{chan2016large}
K.~C. Chan, Z.-H. You, et~al.
\newblock Large-scale prediction of drug-target interactions from deep
  representations.
\newblock In {\em Neural Networks (IJCNN), 2016 International Joint Conference
  on}, pages 1236--1243. IEEE, 2016.

\bibitem{chen2016xgboost}
T.~Chen and C.~Guestrin.
\newblock Xgboost: A scalable tree boosting system.
\newblock In {\em Proceedings of the 22nd acm sigkdd international conference
  on knowledge discovery and data mining}, pages 785--794. ACM, 2016.

\bibitem{chen2015higgs}
T.~Chen and T.~He.
\newblock Higgs boson discovery with boosted trees.
\newblock In {\em NIPS 2014 Workshop on High-energy Physics and Machine
  Learning}, pages 69--80, 2015.

\bibitem{chetlur2014cudnn}
S.~Chetlur, C.~Woolley, P.~Vandermersch, J.~Cohen, J.~Tran, B.~Catanzaro, and
  E.~Shelhamer.
\newblock cudnn: Efficient primitives for deep learning.
\newblock {\em arXiv preprint arXiv:1410.0759}, 2014.

\bibitem{chollet2015keras}
F.~Chollet et~al.
\newblock Keras, 2015.

\bibitem{ciregan2012multi}
D.~Ciregan, U.~Meier, and J.~Schmidhuber.
\newblock Multi-column deep neural networks for image classification.
\newblock In {\em Computer Vision and Pattern Recognition (CVPR), 2012 IEEE
  Conference on}, pages 3642--3649. IEEE, 2012.

\bibitem{cobanoglu2013predict}
M.~C. Cobanoglu, C.~Liu, F.~Hu, Z.~N. Oltvai, and I.~Bahar.
\newblock Predicting drug--target interactions using probabilistic matrix
  factorization.
\newblock {\em Journal of chemical information and modeling},
  53(12):3399--3409, 2013.

\bibitem{dahl2012context}
G.~E. Dahl, D.~Yu, L.~Deng, and A.~Acero.
\newblock Context-dependent pre-trained deep neural networks for
  large-vocabulary speech recognition.
\newblock {\em IEEE Transactions on Audio, Speech, and Language Processing},
  20(1):30--42, 2012.

\bibitem{davis2011comprehensive}
M.~I. Davis, J.~P. Hunt, S.~Herrgard, P.~Ciceri, L.~M. Wodicka, G.~Pallares,
  M.~Hocker, D.~K. Treiber, and P.~P. Zarrinkar.
\newblock Comprehensive analysis of kinase inhibitor selectivity.
\newblock {\em Nature biotechnology}, 29(11):1046--1051, 2011.

\bibitem{donahue2014decaf}
J.~Donahue, Y.~Jia, O.~Vinyals, J.~Hoffman, N.~Zhang, E.~Tzeng, and T.~Darrell.
\newblock Decaf: A deep convolutional activation feature for generic visual
  recognition.
\newblock In {\em ICML}, pages 647--655, 2014.

\bibitem{edwards2011too}
A.~M. Edwards, R.~Isserlin, G.~D. Bader, S.~V. Frye, T.~M. Willson, and H.~Y.
  Frank.
\newblock Too many roads not taken.
\newblock {\em Nature}, 470(7333):163, 2011.

\bibitem{fedorov2010targeted}
O.~Fedorov, S.~M{\"u}ller, and S.~Knapp.
\newblock The (un) targeted cancer kinome.
\newblock {\em Nature chemical biology}, 6(3):166, 2010.

\bibitem{friedman2001greedy}
J.~H. Friedman.
\newblock Greedy function approximation: a gradient boosting machine.
\newblock {\em Annals of statistics}, pages 1189--1232, 2001.

\bibitem{gabel2014beware}
J.~Gabel, J.~Desaphy, and D.~Rognan.
\newblock Beware of machine learning-based scoring functions on the danger of
  developing black boxes.
\newblock {\em Journal of chemical information and modeling},
  54(10):2807--2815, 2014.

\bibitem{gomes2017atomic}
J.~Gomes, B.~Ramsundar, E.~N. Feinberg, and V.~S. Pande.
\newblock Atomic convolutional networks for predicting protein-ligand binding
  affinity.
\newblock {\em arXiv preprint arXiv:1703.10603}, 2017.

\bibitem{gomez2016automatic}
R.~G{\'o}mez-Bombarelli, D.~Duvenaud, J.~M. Hern{\'a}ndez-Lobato,
  J.~Aguilera-Iparraguirre, T.~D. Hirzel, R.~P. Adams, and A.~Aspuru-Guzik.
\newblock Automatic chemical design using a data-driven continuous
  representation of molecules.
\newblock {\em arXiv preprint arXiv:1610.02415}, 2016.

\bibitem{gonen2012predict}
M.~G{\"o}nen.
\newblock Predicting drug--target interactions from chemical and genomic
  kernels using bayesian matrix factorization.
\newblock {\em Bioinformatics}, 28(18):2304--2310, 2012.

\bibitem{gonen2005concordance}
M.~G{\"o}nen and G.~Heller.
\newblock Concordance probability and discriminatory power in proportional
  hazards regression.
\newblock {\em Biometrika}, 92(4):965--970, 2005.

\bibitem{graves2013speech}
A.~Graves, A.-r. Mohamed, and G.~Hinton.
\newblock Speech recognition with deep recurrent neural networks.
\newblock In {\em 2013 IEEE international conference on acoustics, speech and
  signal processing}, pages 6645--6649. IEEE, 2013.

\bibitem{hamanaka2016cgbvs}
M.~Hamanaka, K.~Taneishi, H.~Iwata, J.~Ye, J.~Pei, J.~Hou, and Y.~Okuno.
\newblock Cgbvs-dnn: Prediction of compound-protein interactions based on deep
  learning.
\newblock {\em Molecular Informatics}, 2016.

\bibitem{he2017simboost}
T.~He, M.~Heidemeyer, F.~Ban, A.~Cherkasov, and M.~Ester.
\newblock Simboost: a read-across approach for predicting drug--target binding
  affinities using gradient boosting machines.
\newblock {\em Journal of cheminformatics}, 9(1):24, 2017.

\bibitem{hinton2012deepspeech}
G.~Hinton, L.~Deng, D.~Yu, G.~E. Dahl, A.-r. Mohamed, N.~Jaitly, A.~Senior,
  V.~Vanhoucke, P.~Nguyen, T.~N. Sainath, et~al.
\newblock Deep neural networks for acoustic modeling in speech recognition: The
  shared views of four research groups.
\newblock {\em IEEE Signal Processing Magazine}, 29(6):82--97, 2012.

\bibitem{hochreiter2007fast}
S.~Hochreiter, M.~Heusel, and K.~Obermayer.
\newblock Fast model-based protein homology detection without alignment.
\newblock {\em Bioinformatics}, 23(14):1728--1736, 2007.

\bibitem{jastrzkebski2016learning}
S.~Jastrzkeski, D.~Lesniak, and W.~M. Czarnecki.
\newblock Learning to smile (s).
\newblock {\em arXiv preprint arXiv:1602.06289}, 2016.

\bibitem{kang2014convolutional}
L.~Kang, P.~Ye, Y.~Li, and D.~Doermann.
\newblock Convolutional neural networks for no-reference image quality
  assessment.
\newblock In {\em Proceedings of the IEEE Conference on Computer Vision and
  Pattern Recognition}, pages 1733--1740, 2014.

\bibitem{kimeldorf1971some}
G.~Kimeldorf and G.~Wahba.
\newblock Some results on tchebycheffian spline functions.
\newblock {\em Journal of mathematical analysis and applications},
  33(1):82--95, 1971.

\bibitem{kingma2014adam}
D.~Kingma and J.~Ba.
\newblock Adam: A method for stochastic optimization.
\newblock {\em arXiv preprint arXiv:1412.6980}, 2014.

\bibitem{lecun2015deep}
Y.~LeCun, Y.~Bengio, and G.~Hinton.
\newblock Deep learning.
\newblock {\em Nature}, 521(7553):436--444, 2015.

\bibitem{leung2014deep}
M.~K. Leung, H.~Y. Xiong, L.~J. Lee, and B.~J. Frey.
\newblock Deep learning of the tissue-regulated splicing code.
\newblock {\em Bioinformatics}, 30(12):i121--i129, 2014.

\bibitem{li2015low}
H.~Li, K.-S. Leung, M.-H. Wong, and P.~J. Ballester.
\newblock Low-quality structural and interaction data improves binding affinity
  prediction via random forest.
\newblock {\em Molecules}, 20(6):10947--10962, 2015.

\bibitem{liu2017deep}
X.~Liu.
\newblock Deep recurrent neural network for protein function prediction from
  sequence.
\newblock {\em arXiv preprint arXiv:1701.08318}, 2017.

\bibitem{ma2015deep}
J.~Ma, R.~P. Sheridan, A.~Liaw, G.~E. Dahl, and V.~Svetnik.
\newblock Deep neural nets as a method for quantitative structure--activity
  relationships.
\newblock {\em Journal of chemical information and modeling}, 55(2):263--274,
  2015.

\bibitem{muller2018recurrent}
A.~T. Muller, J.~A. Hiss, and G.~Schneider.
\newblock Recurrent neural network model for constructive peptide design.
\newblock {\em Journal of Chemical Information and Modeling}, 2018.

\bibitem{nair2010rectified}
V.~Nair and G.~E. Hinton.
\newblock Rectified linear units improve restricted boltzmann machines.
\newblock In {\em Proceedings of the 27th international conference on machine
  learning (ICML-10)}, pages 807--814, 2010.

\bibitem{o2016ligand}
M.~J. O’Meara, S.~Ballouz, B.~K. Shoichet, and J.~Gillis.
\newblock Ligand similarity complements sequence, physical interaction, and
  co-expression for gene function prediction.
\newblock {\em PloS one}, 11(7):e0160098, 2016.

\bibitem{oprea2012drug}
T.~Oprea and J.~Mestres.
\newblock Drug repurposing: far beyond new targets for old drugs.
\newblock {\em The AAPS journal}, 14(4):759--763, 2012.

\bibitem{ozturk2016comparative}
H.~{\"O}zt{\"u}rk, E.~Ozkirimli, and A.~{\"O}zg{\"u}r.
\newblock A comparative study of smiles-based compound similarity functions for
  drug-target interaction prediction.
\newblock {\em BMC bioinformatics}, 17(1):128, 2016.

\bibitem{ozturk2018novel}
H.~{\"O}zt{\"u}rk, E.~Ozkirimli, and A.~{\"O}zg{\"u}r.
\newblock A novel methodology on distributed representations of proteins using
  their interacting ligands.
\newblock {\em Bioinformatics, accepted for publication}, 2018.

\bibitem{pahikkala2014toward}
T.~Pahikkala, A.~Airola, S.~Pietil{\"a}, S.~Shakyawar, A.~Szwajda, J.~Tang, and
  T.~Aittokallio.
\newblock Toward more realistic drug--target interaction predictions.
\newblock {\em Briefings in bioinformatics}, page bbu010, 2014.

\bibitem{ragoza2017protein}
M.~Ragoza, J.~Hochuli, E.~Idrobo, J.~Sunseri, and D.~R. Koes.
\newblock Protein--ligand scoring with convolutional neural networks.
\newblock {\em J. Chem. Inf. Model}, 57(4):942--957, 2017.

\bibitem{rose2016rcsb}
P.~W. Rose, A.~Prli{\'c}, A.~Altunkaya, C.~Bi, A.~R. Bradley, C.~H. Christie,
  L.~D. Costanzo, J.~M. Duarte, S.~Dutta, Z.~Feng, et~al.
\newblock The rcsb protein data bank: integrative view of protein, gene and 3d
  structural information.
\newblock {\em Nucleic acids research}, page gkw1000, 2016.

\bibitem{shar2016pred}
P.~A. Shar, W.~Tao, S.~Gao, C.~Huang, B.~Li, W.~Zhang, M.~Shahen, C.~Zheng,
  Y.~Bai, and Y.~Wang.
\newblock Pred-binding: large-scale protein--ligand binding affinity
  prediction.
\newblock {\em Journal of enzyme inhibition and medicinal chemistry},
  31(6):1443--1450, 2016.

\bibitem{simonyan2014very}
K.~Simonyan and A.~Zisserman.
\newblock Very deep convolutional networks for large-scale image recognition.
\newblock {\em arXiv preprint arXiv:1409.1556}, 2014.

\bibitem{smith1981identification}
T.~F. Smith and M.~S. Waterman.
\newblock Identification of common molecular subsequences.
\newblock {\em Journal of molecular biology}, 147(1):195--197, 1981.

\bibitem{srivastava2014dropout}
N.~Srivastava, G.~E. Hinton, A.~Krizhevsky, I.~Sutskever, and R.~Salakhutdinov.
\newblock Dropout: a simple way to prevent neural networks from overfitting.
\newblock {\em Journal of Machine Learning Research}, 15(1):1929--1958, 2014.

\bibitem{tang2014making}
J.~Tang, A.~Szwajda, S.~Shakyawar, T.~Xu, P.~Hintsanen, K.~Wennerberg, and
  T.~Aittokallio.
\newblock Making sense of large-scale kinase inhibitor bioactivity data sets: a
  comparative and integrative analysis.
\newblock {\em Journal of Chemical Information and Modeling}, 54(3):735--743,
  2014.

\bibitem{tian2015boosting}
K.~Tian, M.~Shao, S.~Zhou, and J.~Guan.
\newblock Boosting compound-protein interaction prediction by deep learning.
\newblock In {\em Bioinformatics and Biomedicine (BIBM), 2015 IEEE
  International Conference on}, pages 29--34. IEEE, 2015.

\bibitem{laarhoven2011}
T.~van Laarhoven, S.~B. Nabuurs, and E.~Marchiori.
\newblock Gaussian interaction profile kernels for predicting drug–target
  interaction.
\newblock {\em Bioinformatics}, 2011.

\bibitem{wallach2015atomnet}
I.~Wallach, M.~Dzamba, and A.~Heifets.
\newblock Atomnet: a deep convolutional neural network for bioactivity
  prediction in structure-based drug discovery.
\newblock {\em arXiv preprint arXiv:1510.02855}, 2015.

\bibitem{wang2017computational}
L.~Wang, Z.-H. You, X.~Chen, S.-X. Xia, F.~Liu, X.~Yan, Y.~Zhou, and K.-J.
  Song.
\newblock A computational-based method for predicting drug--target interactions
  by using stacked autoencoder deep neural network.
\newblock {\em Journal of Computational Biology}, 2017.

\bibitem{wen2017deep}
M.~Wen, Z.~Zhang, S.~Niu, H.~Sha, R.~Yang, Y.~Yun, and H.~Lu.
\newblock Deep-learning-based drug--target interaction prediction.
\newblock {\em Journal of Proteome Research}, 16(4):1401--1409, 2017.

\bibitem{xiong2015human}
H.~Y. Xiong, B.~Alipanahi, L.~J. Lee, H.~Bretschneider, D.~Merico, R.~K. Yuen,
  Y.~Hua, S.~Gueroussov, H.~S. Najafabadi, T.~R. Hughes, et~al.
\newblock The human splicing code reveals new insights into the genetic
  determinants of disease.
\newblock {\em Science}, 347(6218):1254806, 2015.

\bibitem{yamanishi2008predict}
Y.~Yamanishi, M.~Araki, A.~Gutteridge, W.~Honda, and M.~Kanehisa.
\newblock Prediction of drug--target interaction networks from the integration
  of chemical and genomic spaces.
\newblock {\em Bioinformatics}, 24(13):i232--i240, 2008.

\end{thebibliography}

\bibliographystyle{abbrv}

\end{document}